\documentclass[10pt,twocolumn,letterpaper]{article}

\usepackage[pagenumbers]{cvpr} 

\usepackage{booktabs}
\usepackage{array}
\usepackage{graphicx}
\usepackage{makecell}
\usepackage{pifont}
\usepackage{threeparttable}
\DeclareUnicodeCharacter{03C0}{\ensuremath{\pi}}
\definecolor{checkgreen}{HTML}{28734A}
\definecolor{crossred}{HTML}{A64040}
\newcommand{\cmark}{\textcolor{checkgreen}{\ding{51}}}
\newcommand{\xmark}{\textcolor{crossred}{\ding{55}}}

\definecolor{cvprblue}{rgb}{0.21,0.49,0.74}
\usepackage[pagebackref,breaklinks,colorlinks,allcolors=cvprblue]{hyperref}

\title{
  \raisebox{-0.15em}{\includegraphics[height=1.5em]{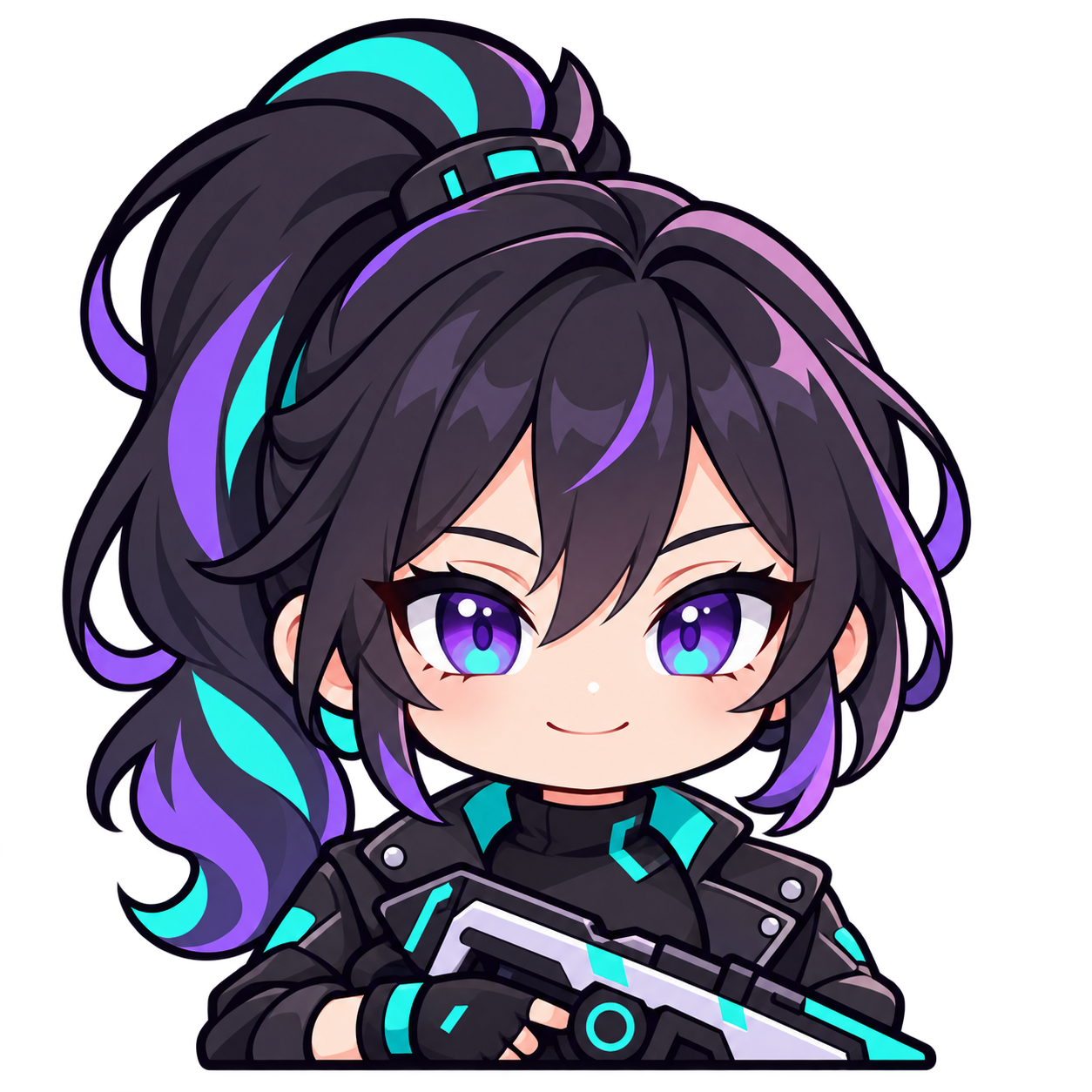}}AceAction-5M: Hierarchical Goals for \\Generalist Action Learning Across 3D Game Worlds
}
\author{
Ruoxuan Zhang$^{1,2*}$ \quad
Kailing Li$^{1,3*}$ \quad
Zheyuan Zhou$^{1}$ \quad
Boyang Chu$^{1}$ \quad
Yingshan Lei$^{1}$ \\[0.3em]
Hongxia Xie$^{2}$ \quad
Liang Du$^{1}$ \quad
Zixun Sun$^{1}$ \\[0.3em]
{\normalsize $^{1}$Tencent} \quad
{\normalsize $^{2}$Jilin University}\quad 
{\normalsize $^{3}$East China Normal University} 
\\[0.3em]
{\tt\small
\href{https://zhangdaxia22.github.io/AceAction-5M/}
{\textcolor{blue}{https://zhangdaxia22.github.io/AceAction-5M/}}
}
}

\begin{document}
\twocolumn[{
    \renewcommand{\twocolumn}[1][]{#1}
    \maketitle

    \begin{center}
        \includegraphics[width=\textwidth]{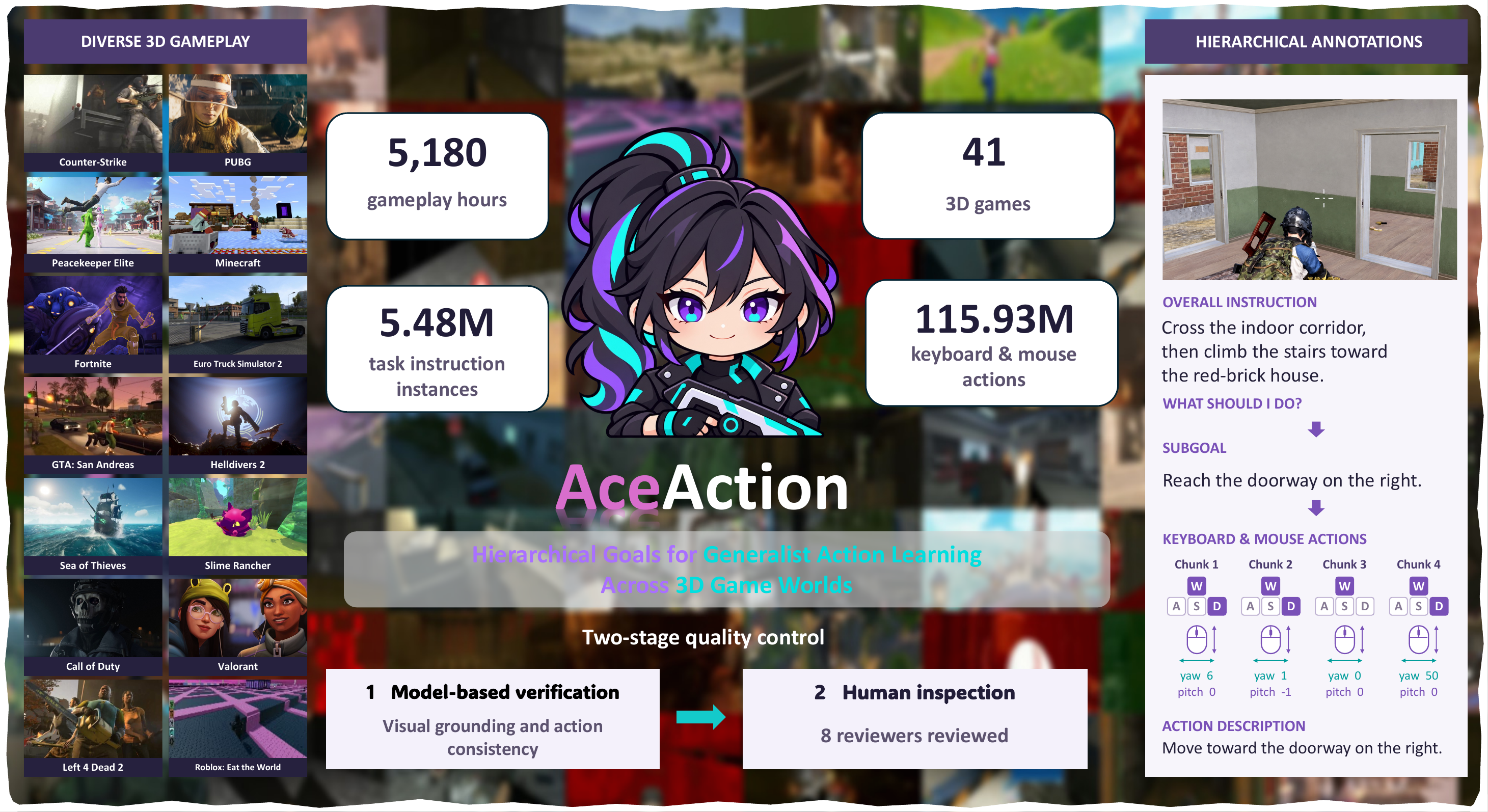}
        \captionof{figure}{\textbf{Overview of the AceAction-5M dataset.}}
        \label{fig:aceaction}
    \end{center}
    \vspace{0.5em}
}]

\begingroup
\renewcommand{\thefootnote}{*}
\footnotetext{Equal contribution.}
\endgroup

\begin{abstract}
Video games provide a practical setting for collecting vision-language-action (VLA) data, whose acquisition in physical environments is costly. Existing datasets, however, remain limited by restricted game coverage, sparse language supervision, or temporal hierarchies that do not necessarily reflect changes in task objectives. We introduce \textbf{AceAction}, a dataset of human gameplay across multiple 3D games for learning skill composition and subgoal transitions. AceAction organizes demonstrations around complete tasks, linking overall instructions, ordered subgoals, and local skill instructions to temporally aligned visual observations and keyboard-and-mouse actions. To construct these annotations, we develop a vision--language model pipeline that identifies subgoals from changes in objectives or interaction targets, anchors navigation goals to visually recognizable waypoints, and composes adjacent subgoals into task instructions. The resulting demonstrations cover navigation, reconnaissance, combat, interaction, and recovery, with explicit stage boundaries that connect task progression to fine-grained control. We assess annotation quality through model-based verification and human inspection. An audit of 500 trajectories yields a 95.8\% pass rate across independent reviewer assessments. AceAction provides a data foundation for studying how agents execute, combine, and switch skills to follow complex instructions in dynamic game environments.
\end{abstract}
    
\section{Introduction}
\label{sec:introduction}
Learning vision-language-action (VLA) models requires demonstrations that connect language goals, visual observations, and executable actions~\cite{pi06,li2026groundingvln,Mao_2026_CVPR,li2026hsgm}. Collecting such data at scale in the physical world is costly, requiring robotic hardware, human supervision, and repeated environment setup~\cite{khazatsky2024droid}. Video games offer a practical setting for collecting large-scale interaction data: human players can perform diverse tasks in rich virtual environments, while visual observations and keyboard-and-mouse actions can be recorded together~\cite{sima2024scaling}. Paired with language annotations, these trajectories support learning to translate task instructions into actions grounded in visual observations. In particular, 3D  games demand close coordination between high-level decision-making and low-level control, requiring agents to navigate, scout, engage opponents, and interact with the environment through rapid and precise keyboard-and-mouse actions. Under partial observability and continuously changing conditions, agents must continually adapt their decisions and actions to environmental feedback.

Tasks in 3D  games often require agents to combine multiple skills across successive stages. For example, an agent asked to \emph{approach the building, scout the entrance, engage the enemy, and fall back to cover to recover} may fight effectively yet continue chasing an opponent when it should retreat. Successful task completion therefore requires agents to understand how each skill serves the current subgoal and to select, compose, and switch skills as the task progresses. This motivates two complementary data requirements. First, \textbf{hierarchical annotations linking tasks, subgoals, and skill-level instructions}: demonstrations should connect overall objectives to local goals and concrete instructions, making explicit what to achieve and how to act, with temporal boundaries indicating subgoal transitions. Second, \textbf{large-scale action data covering diverse goals and tasks}: extensive keyboard-and-mouse trajectories should span varied scenarios and skill combinations, capturing how navigation, reconnaissance, combat, interaction, and recovery are executed and composed for different objectives. Together, these properties support learning both task structure and the behaviors needed to realize it.

Existing game-interaction datasets exhibit limitations in both behavioral coverage and supervision for task structure. EgoCS-400K~\cite{guo2026egocs} provides detailed replay-grounded actions and multi-granularity captions, but remains confined to Counter-Strike and primarily describes local clips and action chains. Open-P2P~\cite{yue2026scaling} broadens game coverage, yet its sparse, goal-oriented instructions provide limited supervision for connecting fine-grained skill execution with intermediate and overall objectives. GameHorizon~\cite{wang2026gamehorizon} introduces multi-horizon instructions aligned with action recordings, but constructs its instruction hierarchy under predefined duration ranges for operations, goals, and strategies. These constraints may split a sustained subgoal or combine several brief subgoals, making temporal hierarchy an imperfect proxy for task-stage structure. These limitations motivate demonstrations with broader behavioral coverage and explicit, objective-driven stage annotations, linking each task to its constituent skills and marking when execution transitions between subgoals.

To bridge this gap, we introduce \textbf{AceAction-5M}, a dataset that unifies multi-game tasks, fine-grained hierarchical instructions, and temporally aligned keyboard-and-mouse trajectories. Each episode pairs a global task description with temporally bounded stage-level subgoals, and aligns local action instructions frame by frame with the keyboard-and-mouse actions that realize them. The resulting trajectories capture how navigation, reconnaissance, combat, interaction, and recovery are composed and switched, providing a data foundation for learning to map complex task instructions to continuous, fine-grained control. Our contributions are as follows:

\begin{itemize}
\item \textbf{A hierarchically annotated, multi-game action dataset.} AceAction covers 42 games and 5,027.87 hours of play, with 5,476,222 episodes in which task descriptions, stage-level subgoals, and local action instructions are jointly aligned with frame-level keyboard-and-mouse trajectories.

\item \textbf{An annotation scheme that makes stage structure explicit.} We define 43 skill categories spanning navigation, reconnaissance, combat, interaction, and recovery, and mark the start and end of every stage, turning \emph{when to switch} from an implicit property of the data into an explicit, learnable signal.
\end{itemize}

\section{Related Work}

\subsection{Game Data and Benchmarks for World Modeling}

Gameplay data supports action-conditioned world modeling by aligning visual observations, actions, and environment states. GameNGen~\cite{valevski2024diffusionmodelsrealtimegame} learns video prediction from DOOM trajectories collected by a reinforcement learning agent, while DIAMOND~\cite{alonso2024diffusionworldmodelingvisual} demonstrates interactive world modeling using recorded CS:GO gameplay. WildWorld~\cite{li2026wildworld} provides frame-level annotations of skeletons, camera poses, and depth for evaluating action following and state consistency.

EgoCS-400K~\cite{guo2026egocs} synchronizes player states, actions, and events extracted from CS:GO/CS2 replays with over 10,000 hours of first-person video. Its multi-granularity captions and video generation prompts link visual changes to replay information, providing language supervision for world modeling. GameVerse~\cite{he2026gameverse} provides minute-scale gameplay videos with temporally coherent camera trajectories and event-aligned structured captions, supporting long-horizon interactive world modeling.

GameHorizon~\cite{wang2026gamehorizon} combines 5,000 hours of gameplay from 21 games with synchronized keyboard-and-mouse actions and three levels of instructions: operations, goals, and strategies. Its offline benchmark evaluates action understanding, goal decomposition, and cross-horizon consistency through multiple-choice questions, with predicted action sequences matched to answer options for scoring. \textbf{Our work} links subtask instructions to waypoints and temporally aligned keyboard-and-mouse trajectories within multi-stage tasks, providing supervision for subgoal transitions and continuous control.

\subsection{Game Action Data for Instruction Following}
\label{sec:related_action_data}

Large-scale gameplay demonstrations support action policy learning~\cite{zhang2026social,zhou2026main}. MineRL~\cite{guss2019minerl} provides human demonstrations in Minecraft with automatically annotated states and actions. VPT~\cite{baker2022video} uses inverse dynamics to infer keyboard-and-mouse actions from Minecraft videos, while NitroGen~\cite{Magne_2026_CVPR} extracts gamepad actions from online videos for cross-game behavior cloning.

Language-based supervision further connects gameplay behaviors to task goals. MineDojo~\cite{fan2022minedojo} combines an open-ended task suite specified in natural language with an internet-scale multimodal knowledge base of videos, tutorials, wiki pages, and forum discussions. Open-P2P~\cite{yue2026scaling} provides cross-game keyboard-and-mouse trajectories with sparse goal-oriented instructions. SIMA~\cite{raad2024scaling} learns to follow language instructions across 3D environments, focusing primarily on short-horizon tasks, while SIMA 2~\cite{bolton2025sima} extends this setting to complex instructions and instruction chaining. STEVE-1~\cite{lifshitz2023steve} supports text- and visual-goal control in Minecraft and composes multi-step behaviors through sequential subgoal prompts.

AceAction organizes demonstrations around complete multi-stage tasks, with annotations spanning overall goals, stage-level subgoals, and local skill instructions. Each subtask is associated with a waypoint, explicit temporal boundaries, and aligned keyboard-and-mouse trajectories. This structure connects individual skill execution to task progression, providing supervision for action execution and the timing of subgoal transitions.

\section{Dataset}
\label{sec:dataset}
\begin{figure*}[t]
    \centering
    \includegraphics[width=\textwidth]{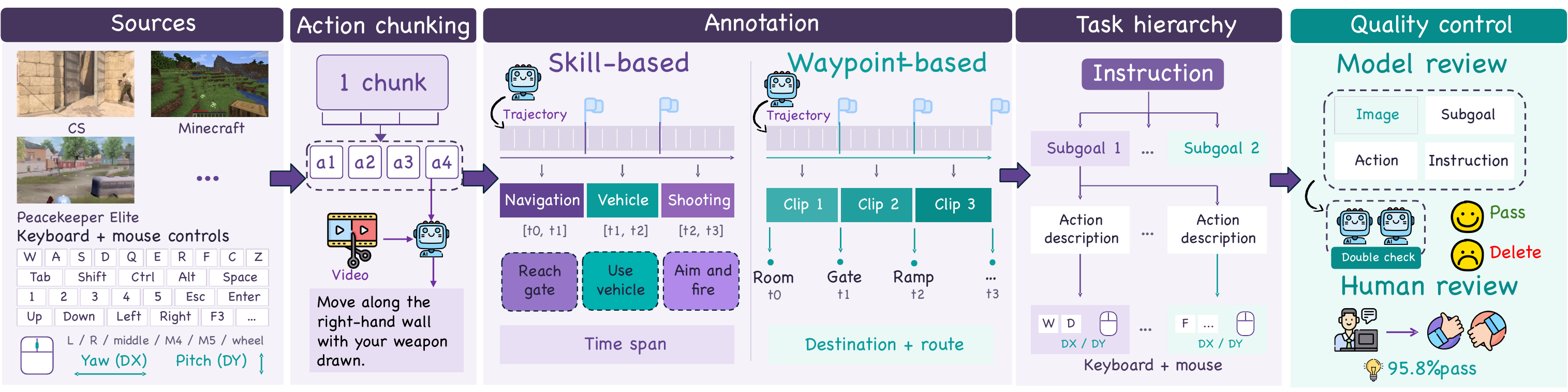}
    \caption{\textbf{Overview of the AceAction data construction pipeline.}
    Gameplay from direct collection and public datasets is annotated with
    skill- and waypoint-based subgoals.
    Adjacent subgoals within each trajectory are composed into task-level
    instructions, followed by annotation and alignment validation.
    Two models then verify visual grounding and action consistency,
    retaining only samples approved by both at every annotation level.
    A sampled human audit further assesses data quality.}
    \label{fig:data_construction}
\end{figure*}
\subsection{Data Source}
Our dataset comprises two complementary components: directly collected gameplay and publicly sourced gameplay. The directly collected gameplay was recorded from full-time in-house staff who held a Gold rank and consented to participate. These recordings were transcribed in anonymized form during data collection. Together, these components provide approximately 5,180 hours of gameplay recordings paired with keyboard-and-mouse actions.

\noindent\textbf{Directly collected gameplay.} We collected gameplay recordings and corresponding keyboard-and-mouse actions from human players. This component includes approximately 1,891 hours of Peacekeeper Elite, including combat scenarios, and 261.17 hours of Valorant. These recordings pair gameplay observations with the actions performed by human players.

\noindent\textbf{Publicly sourced gameplay.} To broaden coverage across games and tasks, we incorporate 2,103.05 hours of CS:GO and CS2 gameplay from match recordings collected online, 811.50 hours of annotated 3D gameplay from Open-P2P, and 113.25 hours of Minecraft gameplay from VPT. These durations refer to the portions included in our dataset rather than the complete upstream datasets.

\noindent\textbf{Action representation and sampling rates.} Across both components, we use a common action representation that encodes keyboard and mouse-button states, together with horizontal and vertical mouse displacement. Our directly collected gameplay is recorded at 16 Hz, corresponding to one action step every 62.5 ms. For publicly sourced gameplay, we CS:GO and CS2 action sequences are 16 Hz, while retaining the original 20 Hz sampling rate for Open-P2P and VPT.

\subsection{Construction Pipeline}
\label{sec:construction}

We construct hierarchical annotations from gameplay observations and actions recorded from human players (Figure~\ref{fig:data_construction}). The pipeline first identifies skill-based subgoals and uses visually recognizable waypoints to construct navigation subgoals. Adjacent subgoals are then composed into overall instructions. Throughout this process, source indices preserve the correspondence between semantic annotations and the original image--action sequences.

\noindent \textbf{Skill-based subgoal annotation.}
We group every four consecutive actions into an annotation unit with a corresponding visual observation. Each trajectory is processed in consecutive windows of up to 32 units. Given the ordered observations and actions, a vision--language model jointly predicts a skill category, start and end indices, and a subgoal description for each segment. The skill category identifies the type of behavior, while the subgoal description specifies its concrete objective. Multiple subgoals can therefore share the same skill category while referring to different targets or objectives.

Segmentation follows continuity of purpose. An activity remains within one segment while it serves the same objective. A new segment begins when the primary objective or interaction target changes. To preserve this continuity across window boundaries, each subsequent window receives the previous window's final observation, skill category, and subgoal description. The model also predicts a continuation flag indicating whether its first segment extends the preceding subgoal. We merge the two segments only when this flag is positive, their skill categories match, and their sequence indices are adjacent. This allows a subgoal to span multiple annotation windows. After normalizing category labels, descriptions, and boundaries, we remove residual airborne segments and map the retained subgoals back to their original image--action sequences using source indices.

\noindent \textbf{Waypoint-based subgoal annotation.}
Navigation subgoals are constructed around arrival events at visually recognizable locations. We identify these events using a sliding window containing up to eight recent observations. The window advances by one observation at each step, and the model evaluates whether the newest frame shows arrival at a specific location. Its input includes the observations, recorded actions, associated skill annotations, previously accepted waypoint names, and the most recent waypoint. Visual observations provide the evidence for arrival, while actions and skill annotations provide behavioral context.

For each accepted arrival event, the model assigns a descriptive waypoint name and associates it with the source frame. Previously accepted names are reused when the same location is recognized again. Continued activity within an already reached location does not produce repeated arrival annotations. Each waypoint thus identifies a destination and the observation at which it is reached.

We then sample clips of varying lengths that end at the waypoint frame. This creates navigation examples with a shared destination but different starting points and intervening actions. For each clip, the model receives sampled keyframes, including the first and arrival frames, together with the complete action sequence, skill context, and target waypoint name. It generates a navigation subgoal that includes the waypoint name and describes the observed route or movement toward that destination. Although annotation uses sampled keyframes, the resulting subgoal retains the complete image--action sequence.
\begin{table*}[t]
    \centering
    \small
    \renewcommand{\arraystretch}{1.25}
    \setlength{\tabcolsep}{4pt}

    \begin{threeparttable}
        \caption{Comparison of gameplay datasets in instruction
        annotations and supervision characteristics.}
        \label{tab:dataset_comparison}

        \begin{tabular*}{0.95\textwidth}{
            @{\extracolsep{\fill}}
            lccc p{0.35\textwidth}@{}
        }
            \toprule
            \thead[l]{Dataset}
            & \thead{Instructions}
            & \thead{Fine-grained hierarchical\\instruction annotations}
            & \thead{Long\\instructions}
            & \thead[l]{Remarks} \\
            \midrule

            Open-P2P~\cite{yue2026scaling}
            & \cmark & \xmark & \xmark
            & Sparse goal-oriented instructions without an explicit
              task--subgoal--skill hierarchy. \\

            EgoCS-400K~\cite{guo2026egocs}
            & \cmark & \xmark & \xmark
            & Restricted to Counter-Strike; replay-grounded actions
              and multi-granularity captions. \\

            CS2-10k~\cite{cs2_10k_2026}
            & \xmark & \xmark & \xmark
            & Restricted to Counter-Strike 2; frame-aligned actions
              without task instructions. \\

            \midrule
            
            \textbf{Ours}
& \cmark & \cmark & \cmark
& \multicolumn{1}{m{0.35\textwidth}@{}}{%
    Task--subgoal--skill instructions with explicit stage
    boundaries and aligned keyboard--mouse actions.%
} \\
            \bottomrule
        \end{tabular*}
    \end{threeparttable}
\end{table*}

\noindent \textbf{Instruction composition.}
We combine adjacent subgoals from the same trajectory in execution order to form an overall instruction. The overall instruction summarizes the task defined by this sequence, while the constituent subgoals retain its intermediate objectives. A composed task can include successive subgoals of the same skill or transitions between different skills. For example, navigation subgoals may lead to a destination where the player subsequently performs an interaction.

For each composed sample, we retain the overall instruction, ordered subgoal descriptions, skill categories, segment boundaries, and corresponding image--action sequence. These boundaries identify the observations and actions associated with each intermediate objective, preserving the relationship between the complete task and its execution stages.

Before export, we check that descriptions are non-empty, skill categories are valid, and segment boundaries are ordered and within the source sequence. We also verify image--action correspondence and confirm that each navigation clip ends at its designated waypoint frame. The resulting samples explicitly connect overall instructions, intermediate subgoals, and executable actions.
\subsection{Quality Control}
\label{sec:quality_control}

We apply a two-stage quality-control procedure consisting of model-based verification and human inspection.

\paragraph{Model-based verification.}
We use Qwen3.8-27B~\cite{qwen38} and GLM-5.3-FLASH~\cite{glm5team2026glm5vibecodingagentic} to review each annotated sample against its gameplay clip and recorded actions. Both models perform bottom-up verification across three annotation levels. First, they check whether each action description matches the visible behavior and recorded controls. Second, they assess whether each subgoal accurately describes the objective supported by its associated observations and actions. Finally, they verify that the overall instruction is consistent with the constituent subgoals and the complete trajectory. At every level, the review checks both visual grounding and action consistency. A sample is retained only when both models approve all annotation levels. A rejection at any level by either model leads to its exclusion.

\paragraph{Human inspection.}
Following model-based verification, eight human reviewers inspect 500 trajectories, with each trajectory assessed by at least two reviewers. The inspection checks whether action descriptions match the observed operations, subgoals reflect the corresponding execution stages, and overall instructions accurately describe the complete tasks. Reviewers assess these annotations against both the gameplay clips and recorded actions. The pass rate across independent review records is \textbf{95.8\%}, where each record represents one reviewer's assessment of one trajectory.

\subsection{Compared with other benchmark}

Table~\ref{tab:dataset_comparison} compares gameplay datasets in terms of instruction annotations and supervision characteristics. Our dataset emphasizes two aspects for training and evaluating game agents, including vision--language--action (VLA) models.
First, trajectory segmentation is grounded in both
player-view videos and recorded keyboard--mouse actions. Each annotated segment preserves the temporal correspondence between visual observations, language instructions, and control inputs, providing aligned supervision for instruction following and action prediction.
Second, our dataset provides extensive subgoal descriptions and second-scale skill annotations within an explicit task--subgoal--skill hierarchy. Overall task instructions are linked to ordered subgoals and action description,
with temporal boundaries marking subgoal transitions.
These annotations provide structured textual supervision for reasoning about task decomposition, skill composition, and subgoal transitions, supporting both training and evaluation on multi-stage tasks.

\section{Conclusion}

We introduce AceAction-5M, a dataset of human gameplay across
multiple 3D games for training and evaluating game agents
and vision--language--action (VLA) models. Combining directly
collected and publicly sourced gameplay, AceAction links
player-view observations and aligned keyboard--mouse actions
through a task--subgoal--skill instruction hierarchy.
Our annotation pipeline identifies subgoals from changes
in objectives or interaction targets, grounds navigation
subgoals in visually recognizable waypoints, and composes
adjacent subgoals into multi-stage task instructions.
Explicit temporal boundaries connect overall task objectives
to local skill execution and mark subgoal transitions.
Model-based verification and human inspection assess
consistency among language annotations, visual observations,
and actions. AceAction provides structured supervision for
studying task decomposition, skill composition, and subgoal
transitions in dynamic game environments.

\clearpage

{
    \small
    \bibliographystyle{ieeenat_fullname}
    \bibliography{main}
}

\end{document}